%% file: main.tex
\documentclass[10pt,twocolumn,letterpaper]{article}

\usepackage[pagenumbers]{cvpr} 

\input{preamble}

\usepackage[many]{tcolorbox}
\definecolor{cvprblue}{rgb}{0.21,0.49,0.74}
\usepackage[pagebackref,breaklinks,colorlinks,allcolors=cvprblue]{hyperref}
\usepackage{pifont}
\usepackage{listings}
\usepackage{enumitem}
\usepackage{tcolorbox}
\usepackage{pdflscape}
\usepackage{rotating}
\usepackage{listings}
\usepackage{xcolor}
\usepackage{adjustbox}

\definecolor{codegreen}{rgb}{0,0.6,0}
\definecolor{codegray}{rgb}{0.5,0.5,0.5}
\definecolor{codepurple}{rgb}{0.58,0,0.82}
\definecolor{backcolour}{rgb}{0.95,0.95,0.92}

\definecolor{darkred}{rgb}{0.5,0,0}
\definecolor{darkgreen}{rgb}{0,0.5,0}
\definecolor{darkblue}{rgb}{0,0,0.55}

\lstdefinestyle{mystyle}{
    backgroundcolor=\color{backcolour},   
    commentstyle=\color{codegreen},
    keywordstyle=\color{magenta},
    numberstyle=\tiny\color{codegray},
    stringstyle=\color{codepurple},
    basicstyle=\ttfamily\scriptsize,
    breakatwhitespace=false,         
    breaklines=true,                 
    captionpos=b,                    
    keepspaces=true,                 
    numbers=left,                    
    numbersep=5pt,                  
    showspaces=false,                
    showstringspaces=false,
    showtabs=false,                  
    tabsize=2
}

\def\paperID{16643} 
\def\confName{CVPR}
\def\confYear{2025}

\title{How to Merge Your Multimodal Models Over \textit{Time}? \raisebox{-0.15\height}{\includegraphics[height=1.4em]{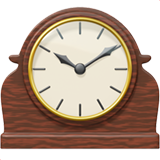}}}

\newcommand{\xmark}{\ding{55}}

\definecolor{DnCBG}{rgb}{0.9, 0.9, 1.}
\usepackage[capitalize]{cleveref}
\crefname{section}{Sec.}{Secs.}
\Crefname{section}{Section}{Sections}
\Crefname{table}{Table}{Tables}
\crefname{table}{Tab.}{Tabs.}

\author{
Sebastian Dziadzio\thanks{\noindent equal contribution, random order $\dagger$equal supervision, random order,\\$\circ$ core contributor. \\Correspondence to: \texttt{vishaal.udandarao}\texttt{@bethgelab.org} or \texttt{sebastian.dziadzio}\texttt{@bethgelab.org}}~~$^{1}$ \quad Vishaal Udandarao$^{\textbf{\scriptsize{*}}1,2}$ \quad Karsten Roth$^{\textbf{\scriptsize{*}}1,3}$ \quad Ameya Prabhu$^{{\circ}1}$ \\ \quad {Zeynep Akata$^{3\dagger}$ \quad Samuel Albanie$^{\dagger}$\quad Matthias Bethge$^{1\dagger}$}\\
\small{
$^1$T\"ubingen AI Center, University of T\"ubingen
$^2$University of Cambridge
$^3$Munich Center for ML, Technical University of Munich}
}

\begin{document}
\maketitle
\input{sec/0_abstract}    
\input{sec/1_intro}
\input{sec/2_related_works}
\input{sec/3_methods}
\input{sec/4_experiments}
\input{sec/5_conclusion}
{
    \small
    \bibliographystyle{ieeenat_fullname}
    \bibliography{main}
}

\appendix
\onecolumn
\input{sec/X_suppl}

\end{document}

%% file: preamble.tex
\definecolor{orange}{RGB}{198, 101, 38}
\definecolor{green}{RGB}{70, 156, 118}
\definecolor{blue}{RGB}{48, 112, 173}
\definecolor{pink}{RGB}{193, 125, 165}
\definecolor{olive}{RGB}{153, 153, 68}
\definecolor{purple}{RGB}{128, 35, 188}
\definecolor{red}{RGB}{188, 45, 45}
\definecolor{yellow}{RGB}{230, 190, 40}
\definecolor{teal}{RGB}{41, 147, 151}

%% file: sec/0_abstract.tex
\begin{abstract}
Model merging combines multiple ``expert'' models---finetuned from a base foundation model on diverse tasks and domains---into a single, more capable model. However, most existing model merging approaches assume that all experts are available simultaneously. In reality, new tasks and domains emerge progressively over time, requiring strategies to integrate the knowledge of expert models as they become available: a process we call \textit{temporal model merging}. The temporal dimension introduces unique challenges not addressed in prior work, raising new questions such as: when training for a new task, should the expert model start from the merged past experts or from the original base model? Should we merge all models at each time step?~Which merging techniques are best suited for temporal merging? Should different strategies be used to initialize the training and deploy the model? To answer these questions, we propose a unified framework called \textsc{TIME}---\underline{T}emporal \underline{I}ntegration of \underline{M}odel \underline{E}xpertise---which defines temporal model merging across three axes: (1) Initialization Phase, (2) Deployment Phase, and (3) Merging Technique. Using \textsc{TIME}, we study temporal model merging across model sizes, compute budgets, and learning horizons on the FoMo-in-Flux benchmark. Our comprehensive suite of experiments across \textsc{TIME} allows us to uncover key insights for temporal model merging, offering a better understanding of current challenges and best practices for effective temporal model merging. Our code is available \href{https://github.com/ExplainableML/fomo_in_flux}{here}.
\end{abstract}

%% file: sec/1_intro.tex
\section{Introduction}\label{sec:intro}
\input{figures/teaser}
Foundation models consolidate a wide range of capabilities and knowledge into a single, large model~\cite{bommasani2023foundation}. Consequently, model merging~\cite{regents1996merge_first,wortsman2022wiseft,ilharco2022patching, li2022branch} has emerged as a key technique for unifying multiple task-specific fine-tuned models derived from a shared base into a single model, effectively preserving the strengths of each specialized expert.

Current model merging approaches typically assume a fixed base model that is fine-tuned independently on $k$ diverse tasks and domains to produce a set of \textit{independent} experts~\cite{garipov2018linearmode,rofin2022linearinterpolationparameterspace,ilharco2022patching,yadav2023tiesmerging,li2022branch}, which are then merged simultaneously. Research in this field has therefore focused primarily on improving merging techniques for larger or structurally differing $k$-sets, exploring the impact of the diversity and scale of finetuning domains, tasks and experts~\cite{wortsman2022modelsoups,yadav2023tiesmerging,sanyal2024lawa,jang2024modelstock}.\vspace{0.1cm}

\noindent
However, the world is constantly evolving, leading to continuous shifts over data distributions, domains, and tasks, with new concepts emerging~\cite{menon2023visual,roth2024practitioner} that may have been insufficiently covered during large-scale pretraining~\cite{koh2021wilds, hu2022lora, pratt2023platypus, udandarao2024active, menon2023visual, udandarao2023sus, roth2023waffle, zhang2021tip, gui2024knn, roth2024context, udandarao2024zeroshot}.
This dynamic nature of real-world applications motivates a hitherto missing systematic exploration into \textit{temporal model merging} (see~\cref{fig:teaser}) to better understand model merging along an additional, overlooked \citep{zhou2024atm,don2022cold} axis: \textit{time}.  
Specifically, in this work, we ask:
\textbf{(1)} Is temporal model merging significantly influenced by the choice of initialization for expert training?
\textbf{(2)} When evaluated over time, which model merging techniques emerge as most suitable?
\textbf{(3)} Should merging strategies differ between initialization and model deployment?
To answer these questions, we propose a unified framework for studying temporal model merging---\textsc{TIME} (\underline{T}emporal \underline{I}ntegration of \underline{M}odel \underline{E}xpertise)---structured around three key axes spanning the design space of temporal merging solutions (as shown in \cref{fig:conmerge-schematic}):\vspace{0.1em}

\noindent \textbf{1. Initialization Phase.} As expert models are trained and created continuously over time, initialization becomes a crucial design element.\vspace{0.1cm}

\noindent \textbf{2. Deployment Phase.} After training an expert on each task, the next step is to deploy a suitable final model. For temporal model merging, this process has to account for a varying number of past expert models and deployed variants, striking a balance between retaining past knowledge and incorporating new task-specific knowledge.\vspace{0.1cm}

\noindent \textbf{3. Model Merging Techniques.} Merging literature has explored a range of approaches for simultaneous merging, from simple weight-averaging and interpolation~\cite{regents1996merge_first, stojanovski2022momentum, ilharco2022patching, roth2024practitioner} to more complex methods involving weight and candidate selection and re-weighting~\cite{yadav2023tiesmerging, matena2022fisher, davari2024breadcrumbs, yu2024dare, marczak2024magmax}. To apply these methods temporally, it is crucial to understand how they perform with varying numbers of merge candidates ($k_t$) and shifting distributions.

\noindent
Using our \textsc{TIME} framework, we position existing model merging approaches along each key axis and conduct a systematic study of model merging over time. For this, we leverage the multimodal FoMo-in-Flux~\cite{roth2024practitioner} benchmark which includes 63 datasets with well-documented sequential properties, enabling a thorough investigation of temporal model merging under practical compute constraints, as proposed in~\citet{roth2024practitioner}. Our experiments systematically explore different merging techniques, initialization, and deployment strategies, providing several key insights:

\begin{tcolorbox}[breakable,width=\linewidth, colback=cvprblue!10, colframe=cvprblue!80!black, title={Key Insights for Temporal Model Merging}]
\noindent
\textcolor{cvprblue}{\textbf{[A] Accounting for time is crucial.}} Standard ``\textit{offline}'' model merging techniques do not generalize well to the temporal merging setting (\cref{subsec:exp_offline}).\vspace{0.1cm}

\noindent
\textcolor{cvprblue}{\textbf{[B] Complex merging techniques matter little.}} Choosing sophisticated merging techniques beyond simple weighted averaging provides at best marginal benefits for temporal model merging, especially for long task sequences (\cref{subsec:exp_online}).\vspace{0.1cm}

\noindent
\textcolor{cvprblue}{\textbf{[C] Initialization and deployment are critical.}} Choosing how to select and combine available weights before and after each task $t$ is most important for temporal model merging (\cref{subsec:framework}).\vspace{0.1cm}

\noindent
\textcolor{cvprblue}{\textbf{[D] Temporal model merging scales well.}} Larger models or compute budgets allows greater benefits from temporal merging. Scaling enables temporal model merging to even outperform the multitask model, trained on all tasks at once (\cref{subsec:scale}).
\end{tcolorbox}

%% file: figures/teaser.tex
\begin{figure}[t!]
    \centering
    \includegraphics[width=\linewidth]{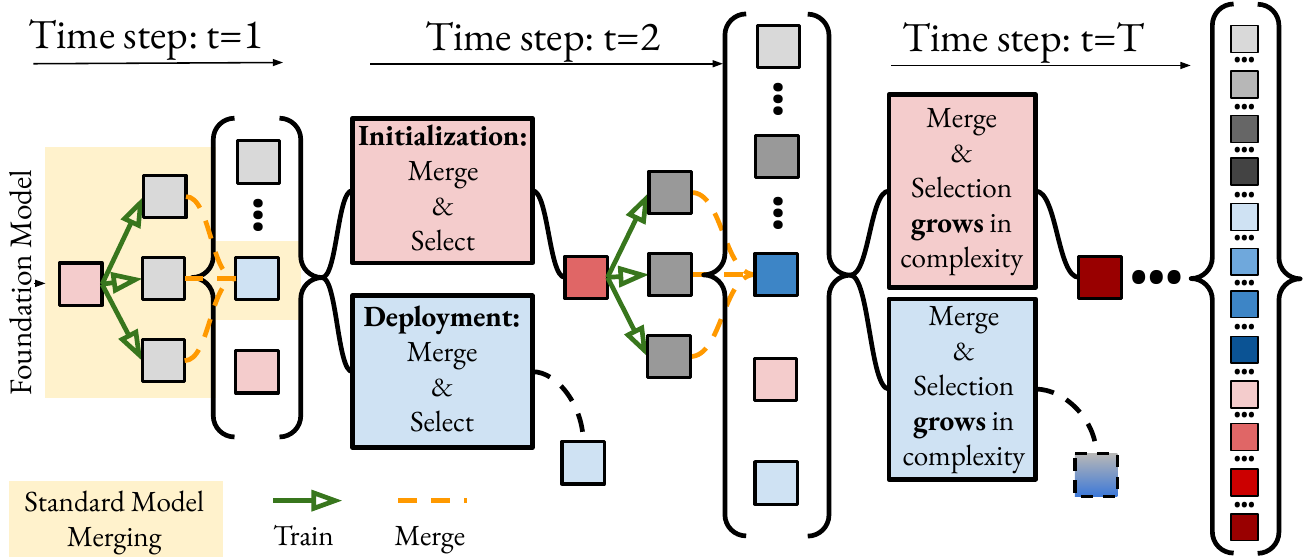}
    \caption{\textbf{Temporal Model Merging} generalizes standard model merging (yellow), which merges multiple trained experts just once, in a single step.
    Our systematic study of this realistic multi-step regime reveals that
    initialization and deployment strategies 
    dominate the importance of the single-step weight merging strategy.
    }
    \vspace{-10pt}
    \label{fig:teaser}
\end{figure}

%% file: sec/2_related_works.tex
\section{Related Works}
\label{sec:rw}

\noindent\textbf{Model Merging.} We provide a short overview of the model merging literature, detailed in these excellent surveys \citep{yadav2024survey, yang2024model}. While both model aggregation through distillation~\cite{roth2024fantastic, cideron2024diversityrewardedcfgdistillation} and averaging checkpoints during training~\citep{kaddour2022stop, sanyal2024lawa, li2024learning} have shown success, the requirement of additional compute limits practicability of these methods \citep{prabhu2023computationally}. Instead, recent work \citep{wortsman2022wiseft, wortsman2022modelsoups, ilharco2022patching, ilharco2023task, rame2023ratatouille, sanyal2024lawa, sung2023empirical, pari2024collective, nylund2023time, zaman2023fuse, stoica2024model, wang2024localizing, he2024localize, oh2024dawin, shen2024efficient, sharma2024non, goddard2024arcee, yadav2024survey, xiong2024multi, yang2024representation, lu2024merge, zheng2024free, nasery2024pleas} has shown the effectiveness of training-free weight averaging and interpolation of fine-tuned expert models to produce an improved base model, benefiting from (linear) mode connectivity in models fine-tuned from a single pre-trained checkpoint~\cite{izmailov2018mode_1,rame2022mode_2,neyshabur2020mode_3,frankle2020mode_4,ainsworth2023mode_5}. These insights have been extended into weight-averaged reward models~\cite{rame2024warm}, policy models~\cite{rame2024warp} with spherical interpolation, and KL-constrained RLHF~\cite{lin2024rhlf_1,liu2024rlhf_2,munos2024rlhf_3,gorbatovski2024rlhf_4}.
Works such as Fisher-Merge~\cite{matena2022fisher}, TIES~\cite{yadav2023tiesmerging}, RegMean~\cite{jin2023regmean}, MATS~\cite{tam2024mats}, DELLA~\cite{deep2024dellamerging}, DARE~\cite{yu2024dare}, Breadcrumbs~\cite{davari2024breadcrumbs}, evolutionary merging~\cite{akiba2024evolutionaryoptimizationmodelmerging} and MagMax~\cite{marczak2024magmax} have explored merging strategies beyond simple interpolation to determine which weights should be merged across expert models. These methods have different benefits for in- and out-of-distribution generalization across domains~\cite{tam2024realisticevaluationmodelmerging}, though recently they have been shown to perform similarly at scale~\cite{yadav2024mattersmodelmergingscale}. Additionally, some works have explored the initialization dimension for effectively merging models \citep{choshen2022start, don2022cold, zhou2024atm, marczak2024magmax}. In this work, we propose a unifying framework for temporal merging and conduct the most comprehensive study of this topic to date.\vspace{0.1cm}

\noindent\textbf{Continual Pretraining} extends beyond standard Continual Learning~\citep{prabhu2023online,roth2024practitioner}, focusing on large-scale model updates starting from pretrained foundation models~\cite{ibrahim2024simple,Garg2023TiCCLIPCT,roth2024practitioner, gui2024knn, prabhu2023categories} and addressing more complex and substantial update tasks~\cite{lin2021clear,cai2021online,liska2022streamingqa,Garg2023TiCCLIPCT,bornschein2023nevis,roth2024practitioner}. There has been limited exploration into using model merging for continual pretraining \citep{marczak2024magmax,alexandrov2024mitigating,stojanovski2022momentum,roth2024practitioner}, as most prior works focus on training strategies including regularization objectives and learning-rate schedules~\citep{roth2024practitioner,prabhu2023computationally,Garg2023TiCCLIPCT,ibrahim2024simple,srivastava2024improving,li2024tic,yildiz2024investigating,thede2024reflecting,ostapenko2022continual,mendieta2023towards}. We keep the training strategy fixed, and provide an in-depth exploration beyond simple merging techniques.

%% file: sec/3_methods.tex
\section{Design Space of Temporal Model Merging}
\label{sec:method}
\input{figures/setup}
\textbf{Notation.} Throughout this work, we use $t$ to refer to a given task at time $t$. Full model parameterization is denoted by $\theta$, with the following key instantiations: $\theta_t$ represents model weights at task $t$, while $\theta_t^I$, $\theta_t^S$, and $\theta_t^O$ denote weights used for \textit{initialization}, \textit{saved} weight checkpoints at task $t$ (i.e. the trained expert models), and the \textit{output} deployed model weights, respectively. Note that while standard model merging considers model weights as elements of a fixed set $\{\theta_k\}_{k=1}^K$, temporal model merging organizes them along the time axis $\theta_t$.

\subsection{Temporal Model Merging through \textsc{TIME}}
Standard model merging is typically performed offline, after all experts have been trained to convergence~\citep{yadav2023tiesmerging,yadav2024mattersmodelmergingscale,gama2014survey,wortsman2022wiseft}. In contrast, model merging in continual pretraining is generally done sequentially, using past checkpoints~\cite{stojanovski2022momentum,roth2024practitioner}. Both approaches are specific instances of our more general temporal merging framework, \textsc{TIME}, which defines temporal merging along \textit{three key axes}: initialization of each expert, merging for deployment at step $t$, and merging techniques $f_\text{merge}$ applied over time:

\subsubsection*{Axis 1: Initialization}
As expert models are created continuously over time, initialization becomes a crucial choice. Unlike model merging at a single point in time, the number of potential starting points grows exponentially over time as new experts are created. This raises the question: should starting points for each time step be derived from the base weights (as in traditional merging), from a merged combination of previous experts, or from most recent weights, as commonly done in continual pretraining~\cite{stojanovski2022momentum, roth2024practitioner}? In this work, we study the following initialization protocols at time step $t$ for \textsc{TIME}:
\begin{itemize}
    \item $\texttt{init}_\text{ZS}$, which consistently initializes with the base zero-shot model weights $\theta_0$ at each timestep $t$.
    \item $\texttt{init}_\text{FT}$, which for step $t$ always initializes with the latest available finetuned model weights $\theta^S_{t-1}$.
    \item $\texttt{init}_\text{EMA}$, which computes an unrolled exponential moving average merge over all previously seen expert models $\{\theta^S_{t'}\}_{1,...,t-1}$ following the equation:
    \begin{equation}\label{eq:ema}
        \theta^\text{EMA}_{t'} = f_\text{merge}\left(\theta^\text{EMA}_{t'-1}, \theta^S_{t'-1}, \mathcal{F}\right)
    \end{equation}
    with merging hyperparameters $\mathcal{F}$. Consequently, the initialization weights are given as $\theta^I_t = \theta^\text{EMA}_{t}$.
\end{itemize}

\subsubsection*{Axis 2: Deployment}
With each update iteration and expert training phase $t$, a decision must be made on the final model to deploy, determining which weights to present for downstream use. In continual pretraining, the trained model $\theta^S_t$ is deployed directly. In contrast, standard model merging applies a merging technique $f_\text{merge}$ to a fixed set of $k$ expert models. Temporal model merging, however, must account for both previously deployed models and the growing number of expert models available over time. Unlike standard merging, where $k$ remains constant, the number of experts to merge increases with each step. As a result, temporal merging introduces the idea of weighted combinations, balancing recent updates with retained past knowledge to achieve adaptability and stability---both critical for effective continual learning~\cite{kirkpatrick2017overcoming,zenke2017continual,prabhu2023computationally,prabhu2023online,roth2024practitioner}. In this work, we study three strategies for model deployment: 
\begin{itemize}
    \item $\texttt{deploy}_\text{FT}$, which at step $t$ deploys the latest finetuned expert model, i.e. $\theta^O_s = \theta^S_t$.
    \item $\texttt{deploy}_\text{EMA}$, which computes an unrolled exponential moving average merge over all expert models, i.e. $\theta^O_t = \theta^\text{EMA}_{t+1}$ following \cref{eq:ema}.
    \item $\texttt{deploy}_\text{ALL}$, which applies a merging technique $f_\text{merge}$ over all previously computed expert models $\{\theta^S_{t'}\}_{1}^{t-1}$.
\end{itemize}

\input{tables/method_comp}

\subsubsection*{Axis 3: Merging Techniques}
At each point in time, for both initialization and deployment, merging technique $f_\text{merge}$ defines \textit{how} to combine the available expert models and checkpoints. In this work, we study nine different variants in total, shown in \cref{tab:merge_methods}. Denoting the number of models to merge at timestep $t$ as $M_t$ (with $t=0$ and $M_t = M$ for standard model merging), we can define these methods as follows:\\

\noindent
\textbf{Weight Averaging~\cite{regents1996merge_first,wortsman2022wiseft,ilharco2022patching,stojanovski2022momentum,roth2024practitioner}} simply employs a uniformly weighted, element-wise average over all models $\theta_{t,i}$, resulting in a merge function $f^\text{WA}_\text{merge}$:
\begin{equation}
    \theta_t = \frac{1}{M_t}\sum_i\theta_{t,i}.
\end{equation}

\noindent
\textbf{SLERP~\cite{shoemake1985AnimatingRW,rame2024warp}} assumes weights to live on a hypersphere, and consequently conducts interpolation along a curved path connecting weight entries. In particular, for two models $\theta_{t,1}$ and $\theta_{t,2}$ deriving from some base weight $\theta_{t-1}$ and the corresponding task vectors~\cite{ilharco2023task} $\delta_{t,i} = \theta_{t,i} - \theta_{t-1}$, SLERP with interrpolation weight $\lambda$ is defined as
\begin{equation}\label{eq:slerp}
    \theta_t = \theta_{t-1} + \frac{\sin (1-\lambda)\Omega_{1,2}}{\sin \Omega_{1,2}}\cdot\delta_{t,1} + \frac{\sin \lambda\Omega_{1,2}}{\sin \Omega_{1,2}}\cdot\delta_{t,2}
\end{equation}
with $\Omega_{1,2}$ being the angle between task vectors $\delta_{t,1}$ and $\delta_{t,2}$. We denote the corresponding merge function $f^\text{SLERP}_\text{merge}$.

\noindent
\textbf{Task Arithmetic~\cite{ilharco2023task}} defines the merge as a function over task vectors $\delta_{t,i} = \theta_{t,i} - \theta_{t-1}$ for each weight $\theta_{t,i}$ fine-tuned from $\theta_{t-1}$. This introduces a simple merge formalism $f^\text{TA}_\text{merge}$ for weighted parameter averaging with a scale $\lambda$:
\begin{equation}
    \theta_t = \theta_{t-1} + \lambda\frac{1}{M_t}\sum_i\delta_{t,i}
\end{equation}\vspace{0.1cm}

\noindent
\textbf{TIES~\cite{yadav2023tiesmerging}} builds on the task arithmetic formalism through controlled pruning of task vector entries with low magnitude. Moreover, the sign for each final merged parameter is set based on the sign of the highest total magnitude across the merge candidates. The final update follows basic task arithmetic, only for entries with matching signs. We refer to the respective merge function as $f^\text{TIES}_\text{merge}$.\vspace{0.1cm}

\noindent
\textbf{DARE~\cite{yu2024dare}} is a similar extension of task arithmetic, but instead of targetted pruning, it randomly zeroes out task vector entries using a random mask $Z_i\sim\text{Bernoulli}(p)$ and masking probability $p$. Final task vector values for $f^\text{DARE}_\text{merge}$ are then rescaled based on $p$:
\begin{equation}
    \delta_{t,i}^\text{DARE} = \frac{(1 - Z_i)\delta_{t,i}}{1-p}.
\end{equation}\vspace{0.1cm}

\noindent
\textbf{Model Stock~\cite{jang2024modelstock}} provides a geometric extension of simple weight averaging as done in Model Soup~\cite{wortsman2022modelsoups} by incorporating base weights $\theta_{t-1}$ into the merging process. Given fine-tuned weights $\theta_{t,1}$ and $\theta_{t,2}$, the Model Stock merge $f^\text{Stock}_\text{merge}$ is defined as follows:
\begin{equation}
    \theta_t = \frac{2\cdot\cos \Omega_{1,2}}{1 + \Omega_{1,2}}\cdot(\theta_{t,2} - \theta_{t,1}) + \left(1 - \frac{2\cdot\cos\Omega_{1,2}}{1 + \cos\Omega_{1,2}}\right),
\end{equation}
utilizing angle $\Omega_{1,2}$ between task vectors $\delta_{t,1}$ and $\delta_{t,2}$.\vspace{0.1cm}

\noindent
\textbf{Breadcrumbs~\cite{davari2024breadcrumbs}} deploys another variation on task arithmetic for model merging. In particular, for a given task vector $\delta_{t,i}$, extreme left and right tails of the absolute magnitude distribution in $\delta_{t,i}$ are zeroed out with left and right thresholds $\beta$ and $\gamma$. The modified task vectors $\delta_{t,i}^\text{Bread}$ are then applied on base weights $\theta_{t-1}$ following the task arithmetic setup, and giving $f^\text{Bread}_\text{merge}$.\vspace{0.1cm}
\input{figures/offline_merging}

\noindent
\textbf{MagMax~\cite{marczak2024magmax}} also uses task vectors---given multiple task vectors $\delta_{t,i}$ (with increments possible along both time $t$ and count axis $i$), the final task vector $\delta_{t}$ is yielded through maximum magnitude entry selection; copying the largest magnitude entries across all $\{\delta_{t,i}\}$ into $\delta_t$, giving $f^\text{Max}_\text{merge}$.\vspace{0.1cm}

\noindent
\textbf{LiNeS~\cite{wang2024linesposttraininglayerscaling}}, for \underline{L}ayer-\underline{i}ncreasing \underline{Ne}twork \underline{S}caling, scales weight updates based on their respective layer depth enabling early layers to remain close to original pretraining weights (cf.~\citet{neyshabur2020mode_3}). Given task vectors $\delta_{t,i}$, now broken down across model layers $\delta^l_{t,i}$ with $l\in[1,...,L]$ and $L$ the number of layers, LiNeS follows the base task arithmetic merging formalism, but updates task vectors as
\begin{equation}
    \delta_{t,i}^\text{LiNeS} = \texttt{concat}\left(\lambda^{l=1}\delta^{l=1}_{t,i},...,\lambda^{l=L}\delta_{t,i}^{l=L}\right)
\end{equation}
with layer-scaled interpolation weights $\lambda^l = \alpha + \beta\frac{l-1}{L-1}$ and hyperparameters $\alpha$, $\beta$, giving $f^\text{Lines}_\text{merge}$.\vspace{0.1cm}

\subsection{Complete Temporal Model Merging Pipeline}
\label{subsec:pipeline}
Incorporating all three axes of temporal model merging, we define a five-stage update pipeline for each task $t$ (see~\cref{fig:conmerge-schematic}), consisting of the following steps:

\begin{enumerate}
    \item \textbf{Init.} Choose one of the aforementioned initialization protocols: $\texttt{init}_\text{ZS}$, $\texttt{init}_\text{FT}$, or $\texttt{init}_\text{EMA}$. This produces initialization weights $\theta^I_t$ at task $t$.
    \item \textbf{Train.} Given $\theta^I_t$, train on current task data $\mathcal{D}_t$ within a set compute budget to produce the expert model $\theta^S_t$.
    \item \textbf{Store.} Append $\theta^S_t$ to storage of expert model weights $\mathcal{S}$.
    \item \textbf{Deploy.} Choose a deployment protocol: $\texttt{deploy}_\text{FT}$, $\texttt{deploy}_\text{EMA}$, or $\texttt{deploy}_\text{ALL}$, and produce the output weights $\theta^O_t=\texttt{deploy}(\mathcal{S})$.
    \item \textbf{Eval.} The deployed $\theta^O_t$ is used for downstream applications and, in our case, extensive evaluation.
\end{enumerate}

\noindent
Particular choices of \texttt{init}, \texttt{deploy} and $f_\text{merge}$ correspond to existing approaches, for example ($\texttt{init}_\text{ZS}, \texttt{deploy}_\text{FT}, f^\text{WA}_\text{merge}$) simply recovers offline merging through weight averaging over experts models derived from original base weights $\theta_0$ for each task $t$. Similarly, ($\texttt{init}_\text{FT}$, $\texttt{deploy}_\text{EMA}$, $f^\text{WA}_\text{merge}$) recovers exponential moving average approaches as done in \cite{stojanovski2022momentum,roth2024practitioner}.

Using our framework, we can explore temporal model merging in a systematic manner across various combinations of all three design choices to understand their impact and recommend best practices.

%% file: figures/setup.tex
\begin{figure*}[!t]
    \centering
    \includegraphics[width=\linewidth]{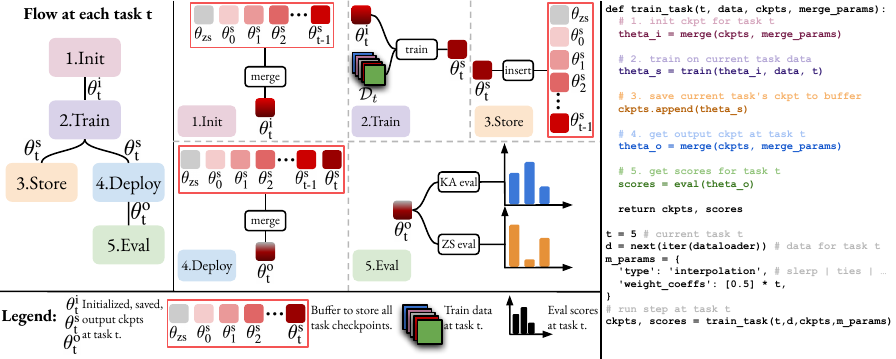}
    \vspace{-18pt}
    \caption{\textbf{Design Space of Temporal Model Merging through \textsc{TIME}.} We showcase our framework for the per-task pipeline of temporal model merging over multiple tasks: At each task $t$, we first initialize the current checkpoint to start training from, $\theta_{t}^{i}$, by using one or more previously stored checkpoints from previous tasks, either directly or by merging them. We train $\theta_{t}^{i}$ on current task data $\mathcal{D}_{t}$ to yield the current task checkpoint $\theta_{t}^{s}$, which is inserted into the checkpoint buffer. Finally, to produce the output model, $\theta_{t}^{o}$, we either merge previously stored checkpoints from the buffer or use them directly. The entire framework is depicted in the pseudo-code on the right panel.}
    \label{fig:conmerge-schematic}
    \vspace{-8pt}
\end{figure*}

%% file: tables/method_comp.tex
\begin{table}
\centering
\caption{\textbf{Comparison of model merging techniques.}}
\resizebox{1\columnwidth}{!}{
\begin{tabular}{lccc}
\toprule
\textbf{Method} & \textbf{Sparsification} & \textbf{Consensus} & \textbf{Scaling} \\
\midrule
Weight averaging~\cite{regents1996merge_first,wortsman2022modelsoups} & \xmark & Linear Int. & Weight coeff. \\
SLERP~\cite{rame2024warp} & \xmark & Spherical Int. & Weight coeff. \\
Task Arithmetic~\cite{ilharco2023task} & \xmark & Linear Int. & Scaling factor \\
MagMax~\cite{marczak2024magmax} & \xmark & Max. Magnitude & Scaling factor \\
TIES~\cite{yadav2023tiesmerging} & Top-k & Sign Agreement & Scaling factor \\
DARE-TIES~\cite{yu2024dare} & Random & Sign Agreement & Scaling factor \\
Breadcrumbs-TIES~\cite{davari2024breadcrumbs} & Top/Bottom-k & Sign Agreement & Scaling factor \\
Model Stock~\cite{jang2024modelstock} & \xmark & Geometric & Adaptive ratio \\
LiNeS~\cite{wang2024linesposttraininglayerscaling} & \xmark & \xmark & Layer weights\\
\bottomrule
\end{tabular}}
\label{tab:merge_methods}
\vspace{-8pt}
\end{table}

%% file: figures/offline_merging.tex
\begin{figure*}[!t]
    \centering
    \includegraphics[width=1\linewidth]{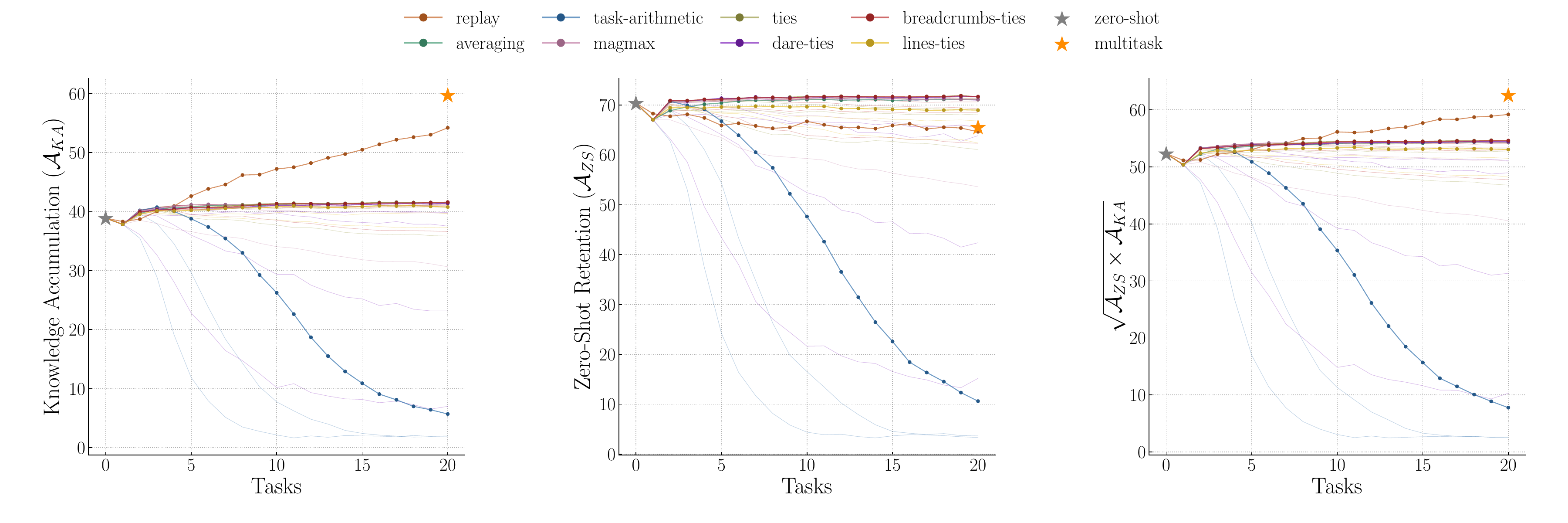}
    \vspace{-27pt}
    \caption{\textbf{Offline merging methods struggle with \textsc{TIME}.} All tested merging techniques perform extremely poorly, and are unable to adapt to the temporal setting, underperforming even a simple \textcolor{orange}{replay} baseline that sequentially trains the base model on task-replayed data.}
    \label{fig:offline_merging}
    \vspace{-10pt}
\end{figure*}

%% file: sec/4_experiments.tex
\section{Experiments}
\label{sec:experiments}

We first discuss experimental details in \cref{subsec:exp_details}. Our initial experiments study the applicability of standard, ``\textit{offline}'' model merging for the temporal paradigm in~\cref{subsec:exp_offline} and possible extensions in~\cref{subsec:gap}. We then transition to a comprehensive exploration of applicable strategies within our \textsc{TIME} framework in~\cref{subsec:framework} and \cref{subsec:exp_online}, before finally scaling up temporal model merging in \cref{subsec:scale}.

\input{figures/gap}

\subsection{Experimental Pipeline Details}\label{subsec:exp_details}
To study temporal model merging in practically relevant scenarios, we use the continual pretraining benchmark Fomo-in-Flux~\cite{roth2024practitioner}. It includes 41 adaptation datasets and 22 separate evaluation datasets, covering an array of visual and semantic distribution shifts, alongside well-defined, practically motivated compute constraints for each task.\vspace{0.1cm}

\noindent
\textbf{Training at task \textit{t}.} By default, we continuously finetune and merge a ViT-B/16 CLIP~\cite{radford2021learning,cherti2022reproducible} model pretrained on LAION-2B~\citep{schuhmann2022laion}. The model is pretrained and finetuned using the standard CLIP objective~\cite{radford2021learning}. Following \cite{roth2024practitioner}, we fix the training steps for each task based on the \texttt{DataComp-Small} computation budget of $1.8\times10^9$ GFLOPS, split equally across 20 tasks. We also additionally explore longer sequences with $T \in \{50, 100\}$. At each temporal merging step, we allow unrestricted access to a pretraining data pool \(\mathcal{P}\), using the same 2M random subset of \texttt{LAION-400M} as in \cite{roth2024practitioner}. We use a cosine-decay LR schedule with a linear warmup of $10\%$, AdamW optimizer~\cite{loshchilov2017decoupled}, a batchsize of 512 and gradient norm clipping to 1. All experiments use PyTorch~\cite{paszke2019pytorch}, and are run on a compute cluster using NVIDIA A100/H100s.\vspace{0.1cm}

\noindent \textbf{Checkpoint Storage.} To enable arbitrary merging strategies at any task $t$, we store each trained expert $\theta^S_t$. We do not store any merged initialization weights $\theta^I_t$ or deployed models $\theta^O_t$, as these can be recovered from stored experts.\vspace{0.1cm}

\noindent \textbf{Evaluation and Metrics.} We focus on two key quantities to evaluate temporal model merging: the level of \textit{adaptation}, reflecting performance improvement with each merging step, and \textit{retention}, capturing the preservation of prior knowledge. Specifically, we report two metrics following \citet{roth2024practitioner}: Knowledge Accumulation (\(\mathcal{A}_{KA}\)), the average accuracy (or recall@5 for retrieval) across all 41 adaptation datasets, and Zero-Shot Retention (\(\mathcal{A}_{ZS}\)), the zero-shot accuracy or recall@5 on all $22$ held-out evaluation datasets. Additional details, including a description of our plotting style, can be found in the supplementary. \vspace{0.1cm}

\subsection{Do We Need Model Merging Across {Time?}}
\label{subsec:exp_offline}

The simplest approach to tackle temporal merging is to disregard the time axis and follow standard \textit{offline} merging paradigms. In \textsc{TIME} terms, this corresponds to a configuration of ($\texttt{init}_\text{ZS}, \texttt{deploy}_\text{ALL}, f_\text{merge}$), which always fine-tunes the initial base weights $\theta_0$. To study the effectiveness of such an offline strategy, we test the above triplet with various choices of $f_\text{merge}$, including $f^\text{WA}_\text{merge}$ (\textcolor{green}{averaging} in \cref{fig:offline_merging}), $f^\text{TA}_\text{merge}$ (\textcolor{blue}{task-arithmetic}), $f^\text{Max}_\text{merge}$ (\textcolor{pink}{magmax}), $f^\text{TIES}_\text{merge}$ (\textcolor{olive}{ties}), $f^\text{DARE}_\text{merge}$ applied over TIES (\textcolor{purple}{dare-ties}), $f^\text{Bread}_\text{merge}$ (\textcolor{red}{breadcrumbs-ties}), and $f^\text{Lines}_\text{merge}$ (\textcolor{yellow}{lines-ties}). To put the results in context, we incorporate \textbf{(1)} a simple continual fine-tuning baseline (\textcolor{orange}{replay}), following \cref{subsec:exp_details}, which replays on both pretraining and previous task data, \textbf{(2)} initial zero-shot ($\theta_0$) performance lower bound, and \textbf{(3)} multitask training upper bound. We visualize trajectories over time for knowledge accumulation $\mathcal{A}_\text{KA}$, zero-shot retention $\mathcal{A}_\text{ZS}$, and the geometric mean of both in \cref{fig:offline_merging}. Our results clearly highlight that there are only marginal differences between merging techniques $f_\text{merge}$ when deployed in an offline manner for a temporal problem, and they all trace similar trajectories in the $\mathcal{A}_\text{KA}$ and $\mathcal{A}_\text{ZS}$ space and achieve similar final performance. We do find that the vanilla $f^\text{TA}_\text{merge}$ (\textcolor{blue}{task-arithmetic}) starts to deviate after a few update steps, an issue fixed by the various subsequent extensions such as TIES, DARE or LiNeS. Overall, however, unlike straightforward continual fine-tuning (\textcolor{orange}{replay}), offline merging with \textit{any} technique fails to address the temporal aspects of the problem, particularly struggling to consistently acquire new knowledge over time (as observed in~\cref{fig:offline_merging}, left).
\input{figures/full_framework}
\begin{tcolorbox}[colback=cvprblue!10, colframe=cvprblue!80!black, title=Key Takeaway]
Offline merging with \textit{any} technique yields similar results, failing to add new knowledge over time.
\end{tcolorbox}

\subsection{Can Replay or Time-Weighting Fix the Issue?}
\label{subsec:gap}
We next ask: ``What are some simple extensions to offline methods to close the gap to the replay baseline''? As the continual fine-tuning baseline \textit{replays} on past data from all previous tasks while training at the current task $t$, can this task data-mixing also help offline merging methods?

\noindent\textbf{Data replaying improves offline merging.} Since offline methods operate entirely under a task-independent assumption, they fail to capture any temporal dependencies. ~\cref{fig:clsoing-the-gap} shows that simply applying data-replay on top of standard offline merging leads to significant boosts in the overall performance. For instance, \textcolor{blue}{best-(offline+replay)} achieves ${58.2}{\%}$ compared to \textcolor{green}{best-offline} at ${54.6}{\%}$, bringing it closer to the \textcolor{orange}{replay} baseline. However, a notable performance gap remains, with \textcolor{blue}{best-(offline+replay)} at ${58.2}{\%}$ falling short of \textcolor{orange}{replay} at ${59.1}{\%}$.

\noindent\textbf{Recency-biased weighting helps.} Next, unlike in standard \textit{offline} averaging, where all task checkpoints are weighted uniformly, we impose temporal ordering via non-uniform weighting for offline merging. We explore several recency-biased, non-uniform weighting schemes, assigning higher weights to more recent tasks to account for the temporal nature of the setting.

We explore various discounting schemes: logarithmic, quadratic, exponential, and cubic, applied to the best offline merge replay method from the previous experiment (please refer to the supplementary for details). As shown in ~\cref{fig:clsoing-the-gap}, these schemes improve performance, with \textcolor{pink}{best-(offline+replay+weighting)} reaching ${58.9}{\%}$, yet still falling slightly short of the \textcolor{orange}{replay} baseline at ${59.1}{\%}$. These results provide strong evidence that accounting for the new temporal axis is crucial for effective temporal model merging, even when implemented as an extension of offline merging.

\begin{tcolorbox}[colback=cvprblue!10, colframe=cvprblue!80!black, title=Key Takeaway]
Accounting for the time aspect is crucial for effective temporal model merging, even as an extension on top of standard offline merging. Still, a small gap to the simple replay baseline remains.
\end{tcolorbox}
\input{figures/online_merging}
\subsection{\textsc{TIME} Travel}
\label{subsec:framework}
In our effort to adapt offline merging to the temporal setting, we have explored two axes of our \textsc{TIME} framework---merging technique and deployment. As discussed in~\cref{subsec:exp_offline}, the choice of merging technique has minimal impact on final performance, while the deployment strategy plays a much more significant role. Next, we systematically explore the design space for temporal merging by testing all valid combinations of three initialization protocols and three deployment protocols described in~\cref{subsec:pipeline}. After discarding incompatible pairs, such as  $\texttt{init}_\text{ZS}$ with $\texttt{deploy}_\text{FT}$, we evaluated the remaining eight variants using weight averaging as the merging technique. As shown in~\cref{fig:full_framework}, the choice of initialization and deployment strategy largely determines performance, significantly affecting both knowledge accumulation and retention.
One combination that stands out consistently is $\texttt{init}_\text{EMA}$ alongside $\texttt{deploy}_\text{EMA}$. This supports the findings of \cite{stojanovski2022momentum,roth2024practitioner} on the applicability of EMA-like solutions to small-scale continual learning and pretraining, while extending the analysis to a much larger, more systematic study of temporal model merging.

As the application of EMA-style model merging achieves a notably better balance between knowledge accumulation and retention than other methods, we call this approach \textit{Best-in-}\textsc{TIME}. In the next section, we will explore the robustness of this strategy across different merging techniques.

\begin{tcolorbox}[colback=cvprblue!10, colframe=cvprblue!80!black, title=Key Takeaway]
The $\text{EMA}$ strategy for initialization and deployment consistently performs the best.
\end{tcolorbox}

\subsection{What is the best merge for Best-in-\textsc{TIME}?}\label{subsec:exp_online}

Having identified the optimal initialization and deployment merging strategy in the previous section using weight-averaging as a canonical merging technique, we now investigate the robustness of our finding by sweeping over other merging techniques, similar to the offline-merging analysis in~\cref{subsec:exp_offline}. In particular, we test $7$ different merging techniques while keeping the \textit{Best-in-}\textsc{TIME} initialization and deployment strategy. From~\cref{fig:online_merging}, it is immediately evident that all merging techniques perform very similarly---in fact, the knowledge accumulation ($\mathcal{A}_{KA}$) of all techniques lie within a single standard deviation of performance variation across multiple runs. This indicates that, over a sufficiently long time horizon, all techniques converge to a similar behavior, echoing our results in~\cref{subsec:exp_offline}. However, we do notice higher variance in the retention metric ($\mathcal{A}_{ZS}$). Overall, our results suggest that for temporal model merging, the specific merging technique used is much less important than selecting the best initialization and deployment strategies.

\begin{tcolorbox}[colback=cvprblue!10, colframe=cvprblue!80!black, title=Key Takeaway]
The choice of merging technique is not particularly important---all techniques perform similarly.
\end{tcolorbox}

\subsection{Scaling Up Temporal Model Merging}\label{subsec:scale}
\input{figures/model_scaling}

We next study the scaling behaviour of temporal model merging across three-axes: \textit{model size}, \textit{compute budget}, and \textit{number of tasks} (results in~\cref{fig:model-scaling} and supplementary). All our experiments utilize the \textit{Best-in-}\textsc{TIME} setup described previously, conducting hyperparameter-optimal EMA-weight-averaging at each task.\vspace{0.1cm}

\noindent
\textbf{Scaling the Model.} As we increase the model scale from $S/16$ (62.3M parameters) to $B/16$ (149.6M), $L/14$ (427.6M), and finally $g/14$ (1.37B) in \cref{fig:model-scaling} (left), we study the tradeoff between new knowledge accumulation and the retention of base knowledge over time. We compare between sequential fine-tuning (circles), and \textit{Best-in-}\textsc{TIME} (squares).
We draw the following conclusions: \textbf{(1)} \textit{Best-in-}\textsc{TIME} scales well with model size, with larger models exhibiting increased affinity to merges over time. This extends and further corroborates standard offline merging insights by \citet{yadav2023tiesmerging}, who showed that model scale facilitates merging efficacy. \textbf{(2)} Moreover, while \citet{roth2024practitioner} highlight better continual fine-tuning with scale, (as also studied by~\citet{ibrahim2024simple}), we show temporal model merging to be substantially more effective across scale. For larger models all the way to the largest ViT-g/14, \textit{Best-in-}\textsc{TIME} vastly outperforms or matches sequential fine-tuning and the multitask target in knowledge retention and positive backward transfer. Moreover, scale facilitates equivalent degrees of knowledge accumulation between sequential fine-tuning and temporal model merging.
Therefore, our model scaling results strongly advocate the use of temporal model merging solutions over standard continual fine-tuning methods~\citep{roth2024practitioner,ibrahim2024simple}.

\noindent
\textbf{Scaling the Compute.} Keeping the underlying base model fixed to ViT-B/16, we next tune the available compute budget similarly to \citet{roth2024practitioner}, and as a result increase the number of allocated update steps per task. We compare a multitask model, trained on all tasks simultaneously, to a budget-optimal \textit{Best-in-}\textsc{TIME}. The only hyperparameter for \textit{Best-in-}\textsc{TIME} is the interpolation weight $w$. For each compute budget, there is a clear optimal choice of that hyperparameter (suboptimal runs shown as gray dots in \cref{fig:model-scaling} (right)). Higher values of $w$ put greater emphasis on accumulation, allowing optimal accumulation-retention trade-offs to be reached at lower compute budgets. However, if a larger compute budget is available, less aggressive temporal model merging can achieve higher absolute trade-offs. Note that in \cref{fig:model-scaling} (right), we report the geometric mean between accumulation and retention, corresponding to the right-most panel in previous plots.

Our results indicate that \textit{Best-in-}\textsc{TIME} scales very well across compute budgets, \textit{clearly approaching the multitask upper bound} in accumulation-retention balance at larger compute budgets. Put together, this gives practitioners the flexibility to adjust \textit{Best-in-}\textsc{TIME} to their compute budget through the choice of $w$. Moreover, with more compute to train each respective expert model, the merging process becomes more effective.

\noindent\textbf{Scaling the Number of Tasks.} Given that all our results until now have been with ${T}{=}{20}$, we next study how \textit{Best-in-}\textsc{TIME} performs as we increase the number of merging time-steps to much longer time-sequences: ${T}{=}{50}$ and ${T}{=}{100}$. From~\cref{fig:full_framework_50,fig:full_framework_100}, we clearly observe that \textit{Best-in-}\textsc{TIME} remains the optimal method of choice across different initialization and deployment strategies.
For ${T}{=}{50}$ and ${T}{=}{100}$, we notice that other methods are initially competitive,but face a significant drop-off in performance near the end of the task sequence, while \textit{best-in-}\textsc{TIME} continues to improve monotonically. This further corroborates that our \textit{Best-in-}\textsc{TIME} is an extremely scalable method for temporal model merging.

\begin{tcolorbox}[colback=cvprblue!10, colframe=cvprblue!80!black, title=Key Takeaway]
Temporal model merging with \textit{Best-in-}\textsc{TIME} scales efficiently across model sizes and the number of tasks. Moreover, compute scaling improves the effectiveness of temporal model merging.
\end{tcolorbox}

%% file: figures/gap.tex
\begin{figure*}[!t]
    \centering
    \includegraphics[width=1\linewidth]{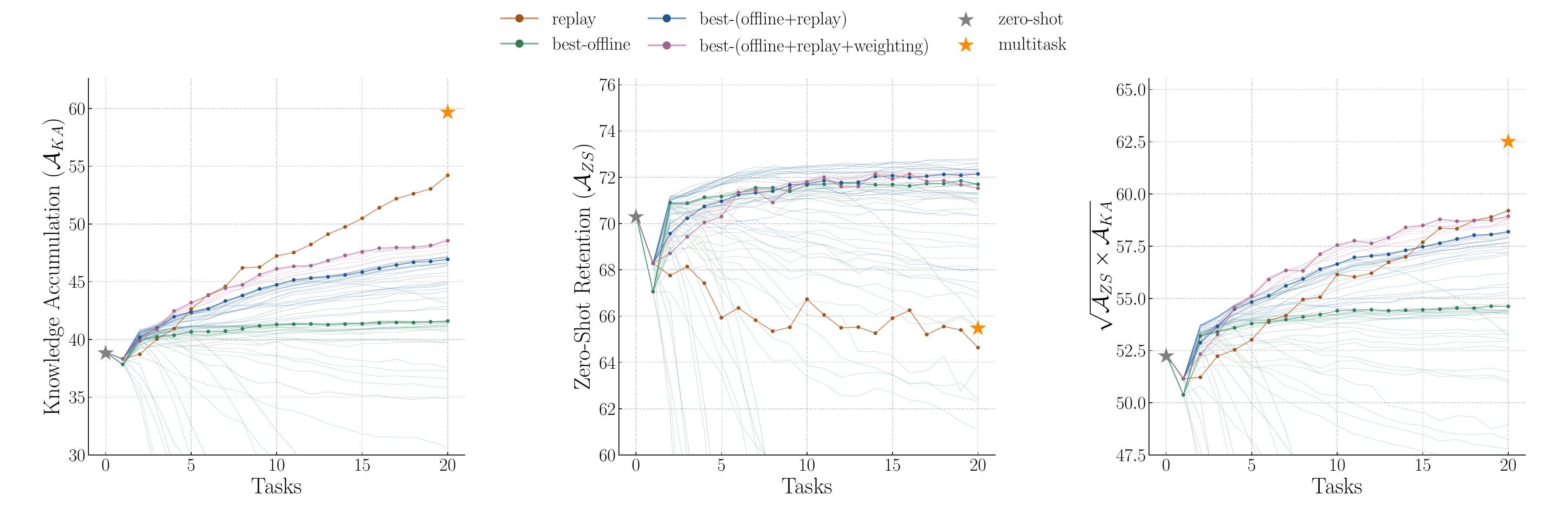}
    \vspace{-27pt}
    \caption{\textbf{Improving \textit{offline} merging.} We identify two simple methods for adapting offline-merging methods to the temporal setting: (1) replaying data from previous tasks (\textcolor{blue}{best-(offline+replay)}) and (2) recency-biased weighting of task checkpoints (\textcolor{pink}{best-(offline+replay+weighting)}). With these method improvements, offline merging methods can match the \textcolor{orange}{replay} baseline.}
    \label{fig:clsoing-the-gap}
    \vspace{-10pt}
\end{figure*}

%% file: figures/full_framework.tex
\begin{figure*}[!t]
    \centering
    \includegraphics[width=1\linewidth]{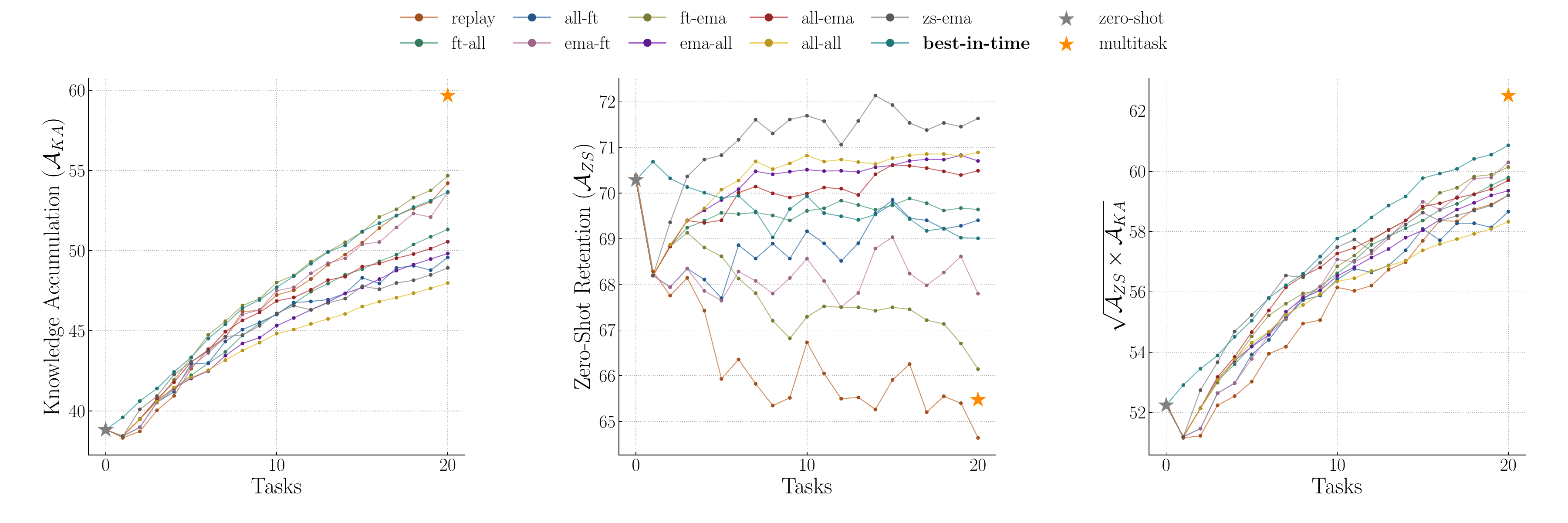}
    \vspace{-27pt}
    \caption{\textbf{A journey through \textsc{TIME}.} We explore various initialization and deployment protocols, finding that the EMA initialization-deployment strikes the best balance between knowledge accumulation and zero-shot retention. We refer to this strategy as \textit{Best-in-}\textsc{TIME}.}
    \label{fig:full_framework}
    \vspace{-10pt}
\end{figure*}

%% file: figures/online_merging.tex
\begin{figure*}[!t]
    \centering
    \includegraphics[width=1\linewidth]{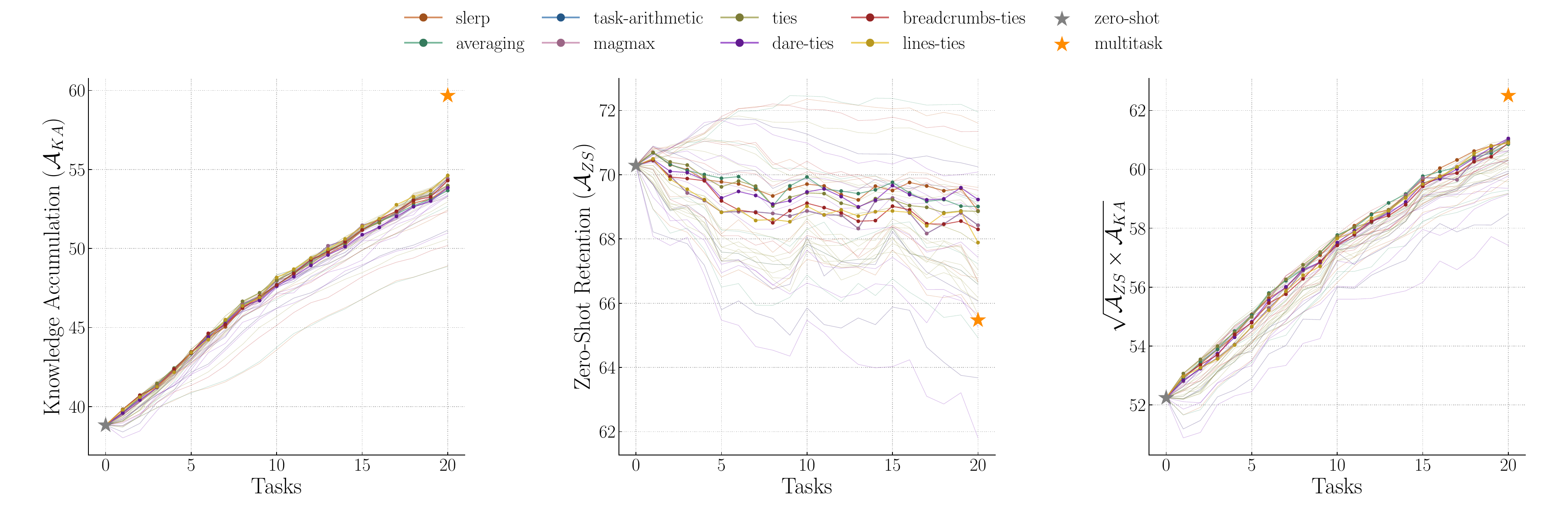}
    \vspace{-27pt}
    \caption{\textbf{Sweeping \textit{Best-in-}\textsc{TIME}.} All merging techniques perform well with the \textit{Best-in-}\textsc{TIME} strategy. Indeed, there are no significant differences between techniques, indicating that initialization and deployment matter more for temporal merging.}
    \label{fig:online_merging}
    \vspace{-10pt}
\end{figure*}

%% file: figures/model_scaling.tex
\begin{figure}
    \centering
    \includegraphics[width=\linewidth]{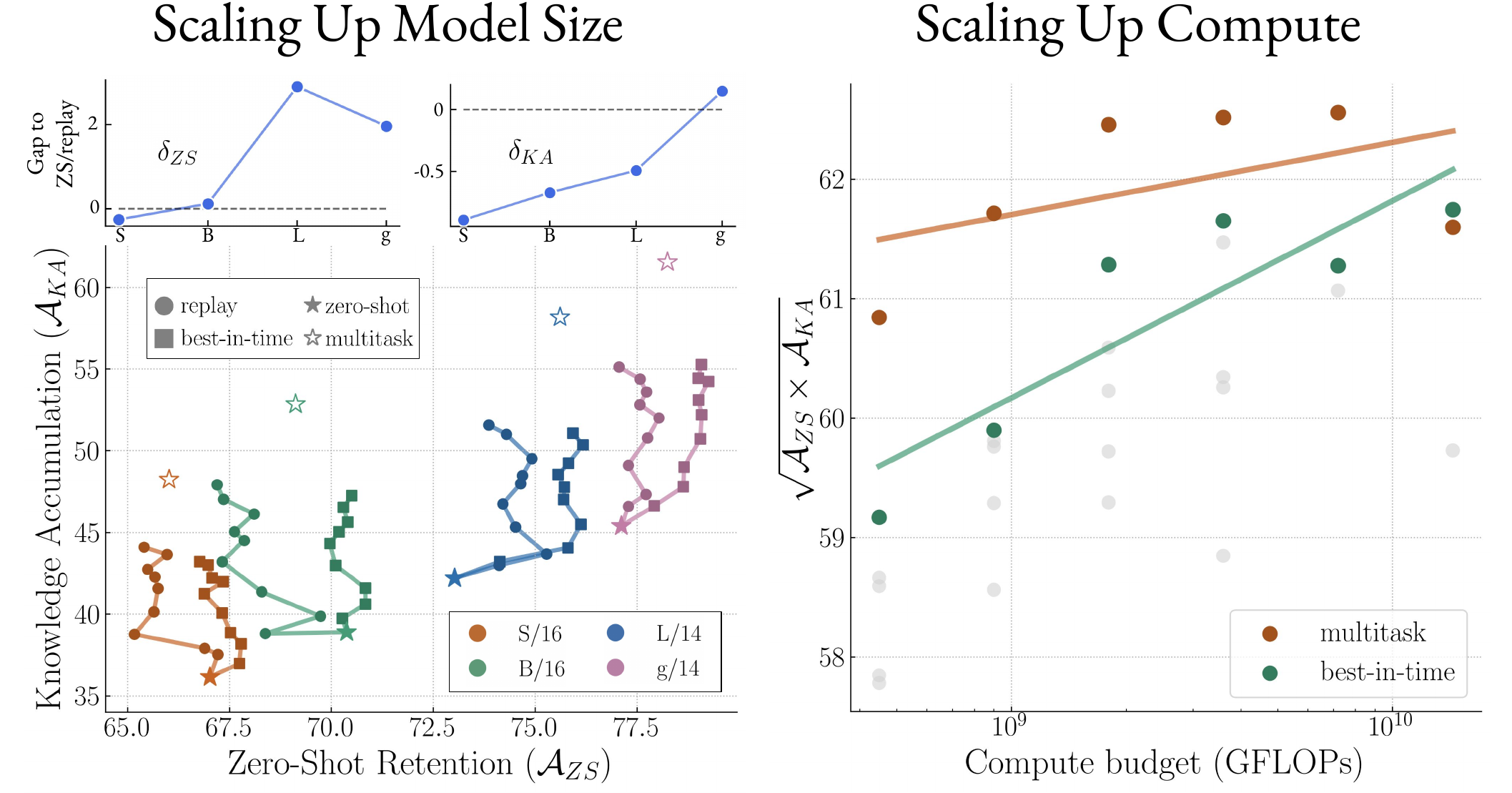}
    \vspace{-10pt}
    \caption{\textbf{Scaling up model merging.} \textit{(left)} With scale, we observe continued improvements of model merging compared to the standard replay baseline.
    \textit{(right)} Our \textit{Best-in-}\textsc{TIME} method continues to improve with scaled total compute budget moving close to the multitask upper-bound. \textcolor{gray}{Gray points} in the plot visualize suboptimal \textit{Best-in-}\textsc{time} hyperparameter-instantiations. 
    }
    \label{fig:model-scaling}
    \vspace{-10pt}
\end{figure}

%% file: sec/5_conclusion.tex
\section{Conclusion}
\label{sec:conclusion}
In this work, we study \textit{temporal model merging}, addressing the challenge of continually merging multimodal models as new tasks and data arrive, and new expert models are trained in succession. To formalize this setting, we propose \textsc{TIME}, a novel unifying framework breaking down temporal model merging ee key axes: (1) initialization phase defining starting weights before each task, (2) deployment phase denoting post-training expert model aggregation, and (3) the choice of merging technique. Using \textsc{TIME}, we conduct a large-scale systematic study uncovering crucial practical guidelines for temporal model merging. Our experiments on the FoMo-in-Flux benchmark spanning $63$ datasets, showcase that accounting for the temporal aspect is crucial, with standard offline merging techniques falling short in this dynamic setting. Moreover, we find the particular choice of merging technique matters far less than the merging strategy for initialization and deployment. Finally, we introduce \textit{Best-in-}\textsc{TIME}, which scales favorably with model size and outperforms existing methods for continual multimodal pretraining. Together, our work provides a systematic entry point into temporal model merging and establishes best practices for this emerging field.

\vspace{0.5cm}
\noindent\textbf{Acknowledgements.} The authors would like to thank (in random order) Shyamgopal Karthik, Shashwat Goel, Ankit Sonthalia, Olivier Hénaff, Alexandre Ramé, and Daniel Marczak for helpful feedback. 
The plotting style in our work is inspired by figures from~\citet{beyer2022knowledge}. The style of~\cref{fig:conmerge-schematic} is inspired by Figure 1 of ~\citet{karamcheti2024prismatic}. VU, KR, and SD thank the International Max Planck Research School for Intelligent Systems (IMPRS-IS). VU, KR, and SD also thank the European Laboratory for Learning and Intelligent Systems (ELLIS) PhD program for support. VU is supported by a Google PhD Fellowship in Machine Intelligence. SA is supported by a Newton Trust Grant. MB acknowledges financial support via the Open Philanthropy Foundation funded by the Good Ventures Foundation. MB is a member of the Machine Learning Cluster of Excellence, funded by the Deutsche Forschungsgemeinschaft (DFG, German Research Foundation) under Germany’s Excellence Strategy – EXC number 2064/1 – Project number 390727645. ZA acknowledges the support from the German Research Foundation (DFG): SFB 1233, Robust Vision: Inference Principles and Neural Mechanisms, project number: 276693517 and ERC Grant DEXIM, project number: 853489.
This research utilized compute resources at the Tübingen Machine Learning Cloud, DFG FKZ INST 37/1057-1 FUGG.

%% file: sec/X_suppl.tex
\clearpage
\section{Plotting Style}
\label{sec:plotting-style-description}

Across \textsc{TIME}, we utilize a common plotting style to visualize our results---with three base subplots (see for \textit{e.g.}, \cref{fig:model-scaling}): 
\begin{itemize}
    \item Knowledge Accumulation ($\mathcal{A}_{KA})$ versus number of tasks over time. In this plot, a gray star indicates the base-weight zero-shot performance on adaptation datasets (see \cref{subsec:exp_details} for more details). An orange star indicates an upper bound achieved through jointly training on all the data at once, with no separation over time.
    \item Zero-Shot Retention ($\mathcal{A}_{ZS}$) versus number of tasks over time. Similar to $\mathcal{A}_{KA}$ versus tasks, this plot visualizes merging results for \textsc{TIME}-variants, but measuring performance on withheld evaluation datasets (cf. \cref{subsec:exp_details}). Again, gray and orange star indicate base and joint training lower and upper bounds, respectively.
    \item Finally, we also aggregate both previous plots into one showcasing the progression of merged performance geometric mean $\sqrt{\mathcal{A}_{ZS} \times \mathcal{A}_{KA}}$ over time; utilizing the same star indication as in the previous subplots.
\end{itemize}
\vspace{0.3cm}
The only deviation from this plotting style is \cref{fig:model-scaling}. The left panel visualizes the trajectory across tasks in the $\mathcal{A}_{KA}$ - $\mathcal{A}_{ZS}$ space. Here, full-colored stars reference base model performance and hollow stars the corresponding joint training upper bounds. The right panel shows the geometric mean of $\mathcal{A}_{KA}$ and $\mathcal{A}_{ZS}$ at the end of the last task for different compute budgets.\\

\noindent
Finally, several plots such as \cref{fig:offline_merging,fig:clsoing-the-gap,fig:online_merging} show the extensive scale of our experiments through background visualizations of sub-optimal hyperparameter choices in lighter colors (as opposed to the optimal choices using darker coloring). This plotting style is loosely inspired by~\citet{beyer2022knowledge}.

\section{Experiments with Tasks as Datasets}
\label{sec:dataset-incremental-results}
In the main text, we presented all results using a data stream that randomly mixes concepts from different datasets into a coherent set of tasks---following the \textit{random} data-stream in~\citet{roth2024practitioner}. Here, we relax this constraint and re-run our experiments using individual datasets as tasks, consistent with the standard model merging literature~\citep{ilharco2022patching,ilharco2023task,yadav2023tiesmerging}. Specifically, we use the \textit{dataset-incremental} stream from~\citet{roth2024practitioner}. Even in this setup, we reproduce our main findings. In~\cref{fig:dataset-incremental-offline}, we confirm the results from~\cref{fig:offline_merging}, showing that all offline merging techniques perform poorly when exposed to the axis of time, failing to even match the performance of a simple continual fine-tuning \textit{replay} baseline. Additionally, in~\cref{fig:dataset-incremental-full-framework}, we corroborate the results from~\cref{fig:full_framework}, demonstrating that the \textit{best-in-}\textsc{TIME} method remains the most effective temporal model merging approach. We also confirm that the choice of model merging technique is far less critical for temporal model merging than the initialization and deployment strategies.

\input{figures/dataset_incremental_offline_merging}
\input{figures/dataset_incremental_full_framework}

\newpage
\section{Experiments with Longer Task Sequences}
\label{sec:longer-task-sequence-results}
\input{figures/full_framework_50}
\input{figures/full_framework_100}
To test the robustness of our findings in \cref{subsec:framework}, we repeat the experiment shown in \cref{fig:full_framework} on a longer sequence with the number of tasks $T=50$ (\cref{fig:full_framework_50}). For 50 tasks, \textit{Best-in-}\textsc{TIME} still strikes the optimal balance between knowledge accumulation and zero-shot retention. One notable difference with respect to \cref{fig:full_framework} is the large initial advantage of the zero-shot initialization strategy combined with the EMA deployment strategy. When the learning horizon is further extended to 100 tasks, this initial advantage is maintained, establishing the zero-shot initialization approach as the best-performing method, as shown in \cref{fig:full_framework_100}. Although the double EMA variant surpasses zero-shot initialization in knowledge accumulation, its poor retention relegates it to third place on the combined metric. In this exploration we re-use the optimal interpolation weight from the 20 task scenario, which may no longer be ideal for longer horizons, as it directly influences the balance between knowledge accumulation and zero-shot retention.

\section{Non-Uniform Weighting Schemes for Improving Offline Merging}

\subsection{Details}

In~\cref{subsec:gap}, we showed that recency-biased non-uniform weighting helps to improve offline merging performance, when used in conjunction with replaying old task data. Typically, when several models are merged using simple weight averaging, they are uniformly averaged. However, this clearly ignores the dimension of time, assuming all previous task-checkpoints as independent and agnostic of time. Hence, we explored 8 non-uniform recency-biased schemes including \textit{linear}, \textit{quadractic}, \textit{sqrt}, \textit{cubic}, \textit{fifth-power}, \textit{tenth-power}, \textit{exponential}, \textit{log}.

\begin{lstlisting}[language=Python, caption=Recency-biased Non-uniform Weighting Algorithms]

'''
Each weighting scheme below produces a list of N values, at each task N. The ith element of the output list denotes the weight coefficient of the ith task checkpoint.
'''

        def linearly_increasing_list(n):
            values = np.linspace(1, n, n)
            return normalize(values)

        def sqrt_scaling_list(n):
            values = np.array([np.sqrt(i) for i in range(1, n + 1)], dtype=float)
            return normalize(values)

        def quadratic_scaling_list(n):
            values = np.array([i**2 for i in range(1, n + 1)], dtype=float)
            return normalize(values)

        def cubic_scaling_list(n):
            values = np.array([i**3 for i in range(1, n + 1)], dtype=float)
            return normalize(values)

        def fifth_power_scaling_list(n):
            values = np.array([i**5 for i in range(1, n + 1)], dtype=float)
            return normalize(values)

        def tenth_power_scaling_list(n):
            values = np.array([i**10 for i in range(1, n + 1)], dtype=float)
            return normalize(values)

        def exponentially_increasing_list(n, base=2):
            values = np.array([base**i for i in range(n)], dtype=float)
            return normalize(values)

        def logarithmic_scaling_list(n):
            values = np.array([np.log(i + 1) for i in range(1, n + 1)], dtype=float)
            return normalize(values)

        def normalize(v):
            v /= v.sum()
            return v.tolist()

\end{lstlisting}

\subsection{Reversed Non-Uniform Weighting Schemes}

\input{figures/reverse_weighting_gap}

In~\cref{fig:clsoing-the-gap}, we found that a simple yet effective method for boosting the performance of offline merging methods is recency-biased non-uniform weighting, i.e. giving larger weights to more recent checkpoints while merging. Here, we ask the question---what if we reversed the weighting schemes such that we give larger weights to older task checkpoints? From~\cref{fig:reverse-streams}, we indeed observe that such a reverse strategy performs worse than the best recency-biased weighting schemes, since the knowledge accumulation ability is hampered by giving more emphasis to older tasks. However, note that such a sub-optimal reverse weighting strategy is still better than the pure offline merging strategy with \textit{no replay}. This helps further ablate the exact importance of \textit{replay} and \textit{non-uniform weighting} for improving pure offline-merging techniques in the presence of the time axis.

\newpage
\section{Variance Analysis across Runs}
\input{figures/run_variance}
To put our results from \cref{subsec:exp_online} in perspective, we quantify the variance across runs for a single merging method. Specifically, we run \textit{Best-in-}\textsc{TIME} three times and show the mean and standard deviation across runs in \cref{fig:run_variance}. Comparing this to \cref{fig:online_merging} reveals that the best results for different methods fall within the standard deviation of multiple runs of the same method. In particular, for the last task, the standard deviation of the geometric mean of knowledge accumulation and zero-shot retention is $0.96$.

\section{Hyperparameter Details}
In an effort to remove any confounding factors, we conduct an extensive hyperparameter sweep, to the best of our abilities, for each individual merging technique for~\cref{fig:offline_merging,fig:clsoing-the-gap,fig:online_merging}. We list the hyperparameter ranges swept over for each technique below:
\begin{itemize}
    \item \textbf{Weight Averaging.} For the offline merging, we use a standard merging coefficient of $\frac{1}{N}$, where $N$ is the number of task checkpoints to merge.
    \item \textbf{SLERP.} In SLERP, as we can only merge two checkpoints at a time, we sweep over the following weight-coefficients: \{0.1,0.3,0.5,0.7,0.9\}.
    \item \textbf{Task-Arithmetic.} We sweep over the scaling factor: \{0.1,0.2,0.3,0.4,0.5,0.6,0.7,0.8,0.9,1.0\}
    \item \textbf{TIES.} We sweep over the scaling factor: \{0.1,0.2,0.3,0.4,0.5,0.6,0.7,0.8,0.9,1.0\} and the pruning-fraction: \{0.1,0.2,0.3,0.4,0.5,0.6,0.7,0.8,0.9,1.0\}.
    \item \textbf{DARE-TIES.} We sweep over the scaling factor: \{0.1,0.2,0.3,0.4,0.5,0.6,0.7,0.8,0.9,1.0\} and the pruning-fraction: \{0.1,0.2,0.3,0.4,0.5,0.6,0.7,0.8,0.9,1.0\}.
    \item \textbf{Breadcrumbs-TIES.} We sweep over the scaling factor: \{0.1,0.2,0.3,0.4,0.5,0.6,0.7,0.8,0.9,1.0\} and the pruning-fraction: \{0.1,0.2,0.3,0.4,0.5,0.6,0.7,0.8,0.9,1.0\}.
    \item \textbf{MagMax.} We sweep over the scaling factor: \{0.2,0.4,0.8,1.0\}.
    \item \textbf{LiNeS-TIES.} We keep $\alpha$ fixed to $0.5$, and sweep $\beta$: \{0.2,0.5,0.8\} and prune-fraction: \{0.2,0.5,0.8\} as recommended in the original paper~\citep{wang2024linesposttraininglayerscaling}.
\end{itemize}

%% file: figures/dataset_incremental_offline_merging.tex
\begin{figure}[h]
    \centering
    \includegraphics[width=\linewidth]{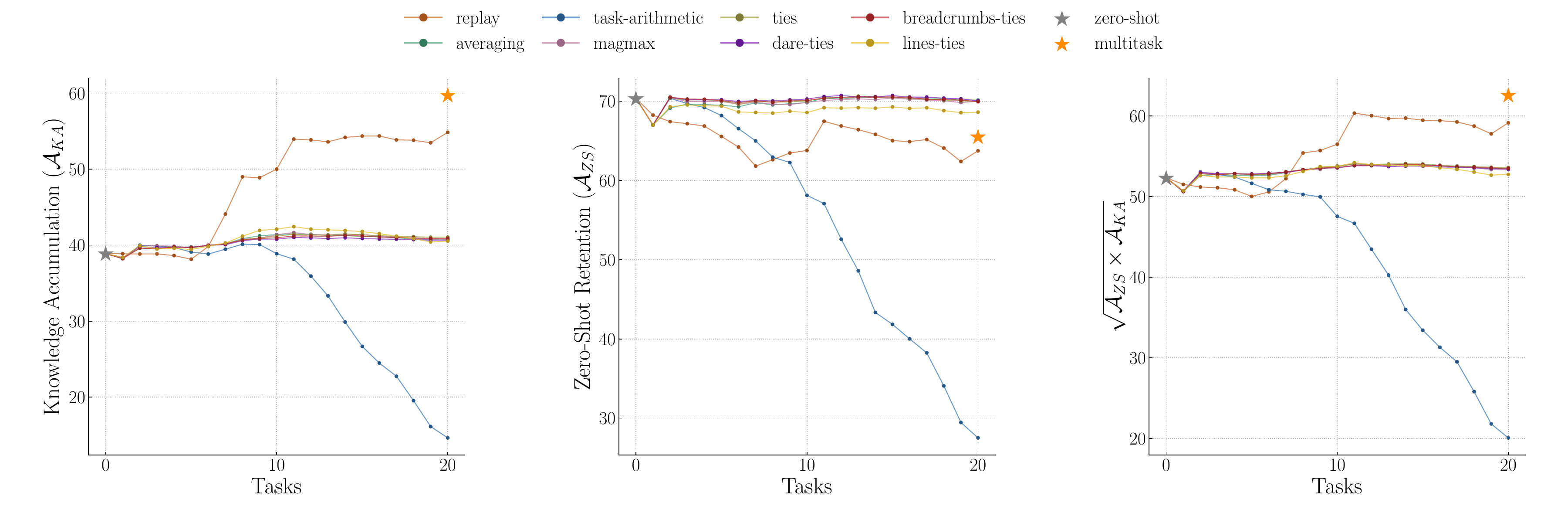}
    \vspace{-20pt}
    \caption{\textbf{Offline merging techniques still struggle in the tasks-as-datasets setting.} Switching from the \textit{random} data-stream (\cref{fig:offline_merging} in the main paper) to the \textit{dataset-incremental} stream, which aligns more closely with the standard multi-task merging literature setups, reveals that offline merging techniques still severely underperform compared to the simple \textit{replay} baseline.}
    \label{fig:dataset-incremental-offline}
\end{figure}

%% file: figures/dataset_incremental_full_framework.tex
\begin{figure}
    \centering
    \includegraphics[width=\linewidth]{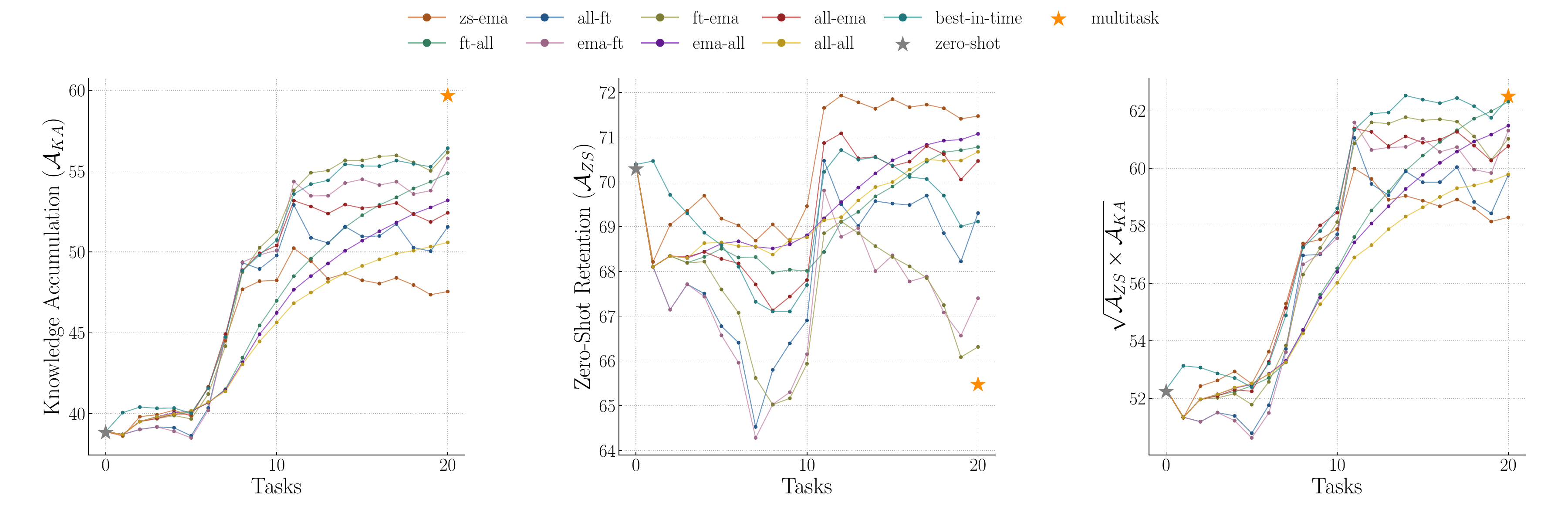}
    \vspace{-20pt}
    \caption{\textbf{Dataset-Incremental TIME Exploration.} We replicate the results from~\cref{fig:full_framework} using the dataset-incremental stream instead of the random stream. The main takeaways remain unchanged: initialization and deployment strategies primarily determine temporal merging performance, and the EMA-averaging initialization and deployment strategy utilized in \textit{Best-in-}\textsc{TIME} is the best approach.}
    \label{fig:dataset-incremental-full-framework}
\end{figure}

%% file: figures/full_framework_50.tex
\begin{figure}[h]
    \centering
    \includegraphics[width=\linewidth]{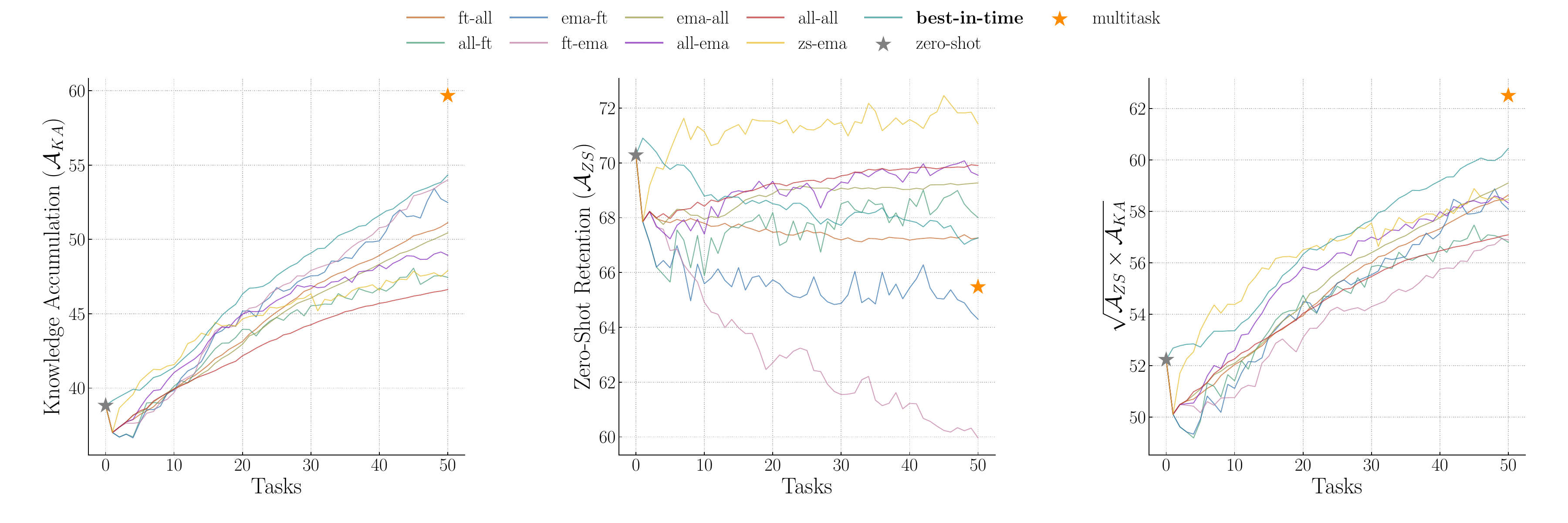}
    \vspace{-20pt}
    \caption{\textbf{A long journey through TIME.} We compare all valid combinations of initialization and deployment protocols on a longer sequence of 50 tasks. \textit{Best-in-}\textsc{TIME} remains the best in balancing knowledge accumulation and zero-shot retention.}
    \label{fig:full_framework_50}
\end{figure}

%% file: figures/full_framework_100.tex
\begin{figure}[h]
    \centering
    \includegraphics[width=\linewidth]{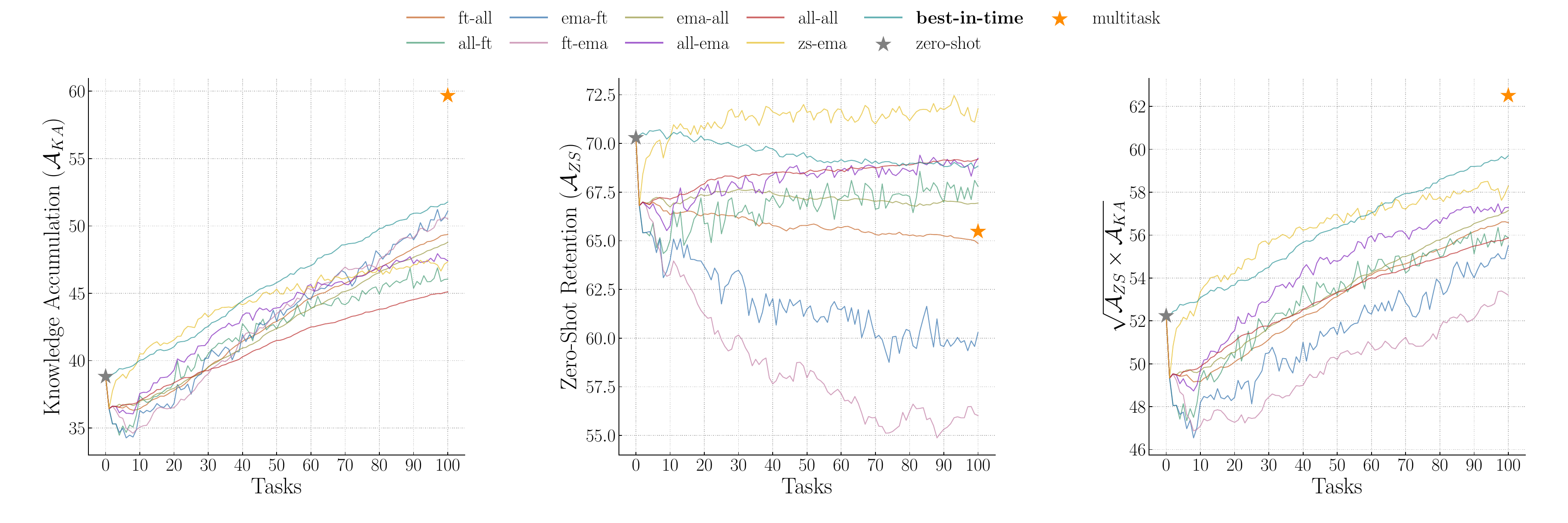}
    \vspace{-20pt}
    \caption{\textbf{An even longer journey through TIME.} We compare all valid combinations of initialization and deployment protocols on a longer sequence of 100 tasks. \textit{Best-in-}\textsc{TIME} still remains the best approach balancing knowledge accumulation and retention, measured as the geometric mean of the two metrics in the right-most figure.}
    \label{fig:full_framework_100}
\end{figure}

%% file: figures/reverse_weighting_gap.tex
\begin{figure*}[h]
    \centering
    \includegraphics[width=1\linewidth]{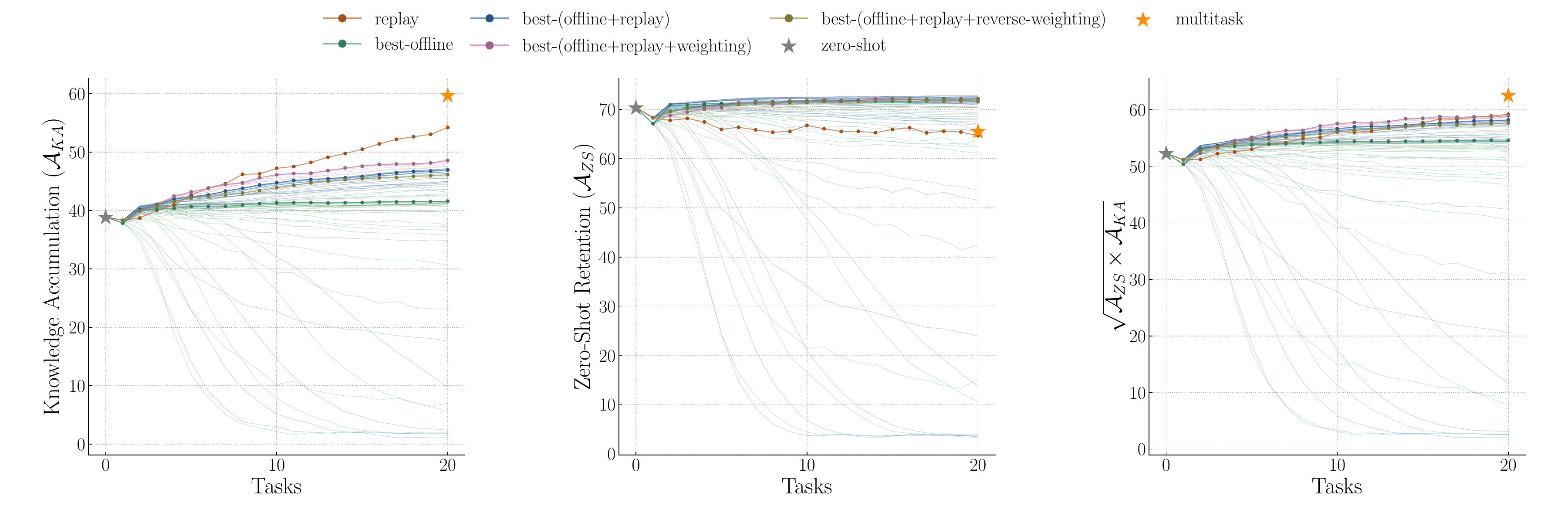}
    \vspace{-20pt}
    \caption{\textbf{Effect of reverse-weighting for offline merging techniques.} We find that reversing the weighting scheme that yielded consistent boosts from~\cref{fig:clsoing-the-gap} is sub-optimal---indeed, it performs worse than the offline merging with replay methods.}
    \label{fig:reverse-streams}
\end{figure*}

%% file: figures/run_variance.tex
\begin{figure}[h]
    \centering
    \includegraphics[width=\linewidth]{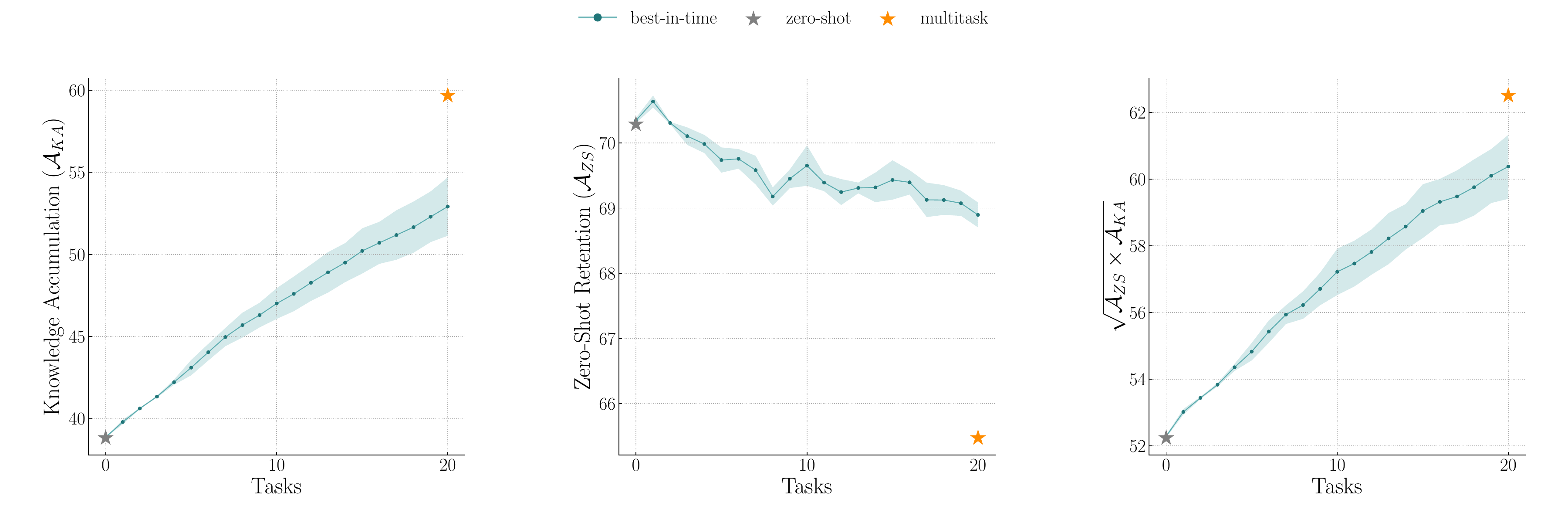}
    \vspace{-20pt}
    \caption{The mean and standard deviation across three runs of \textit{Best-in-}\textsc{TIME}.}
    \label{fig:run_variance}
\end{figure}